\documentclass[twoside,runningheads,10pt,twocolumn]{llncs}

\newcommand{\defaultLlncsTextWidth}[0]{6.5in}
\usepackage{geometry}
\AtBeginDocument{%
  \providecommand\BibTeX{{%
    \normalfont B\kern-0.5em{\scshape i\kern-0.25em b}\kern-0.8em\TeX}}}

\usepackage[disable]{todonotes} 

\usepackage{enumitem}

\usepackage{breakurl} 

\usepackage[breaklinks=true]{hyperref}
\usepackage{breakcites}
\usepackage{bbm}
\usepackage{amsmath}
\usepackage{amssymb}
\usepackage{booktabs}
\usepackage{tabularx}
\usepackage[capitalise]{cleveref} 
\usepackage{multirow}
\usepackage{subfig}
\usepackage{parnotes}
\usepackage{seqsplit}

\usepackage[utf8]{inputenc}
\usepackage[LAE,LFE,T1]{fontenc}
\usepackage[farsi,english]{babel}
\usepackage[linesnumbered,lined,boxed,commentsnumbered]{algorithm2e}
\usepackage{bbm}

\crefname{pseudocodeLine}{line}{lines}
\crefname{pseudocodeLine}{Line}{Lines}

\usepackage{mathrsfs}

\usepackage{amsmath}
\usepackage{amssymb}
\usepackage{comment}

\newcommand{\methNam}{Fanoos}

\newcommand{\abstractState}{abstract state}
\newcommand{\abstractStates}{abstract states}
\newcommand{\debateableSit}{debatable}
\newcommand{\autouserShort}{AU}
\newcommand{\AutouserShort}{AU}
\newcommand{\autouser}{autouser}
\newcommand{\Autouser}{Autouser}
\newcommand{\autouserFull}{automated-user}
\newcommand{\sampleFrom}{\sim}
\newcommand{\FState}{state}
\newcommand{\FStates}{states}
\newcommand{\CapFState}{State}

\newcommand{\SymbFSDescr}{D}
\newcommand{\SymbFState}{q} 
\newcommand{\SymbUQ}{\textbf{quest}} 
\newcommand{\SymbHistory}{H}

\newcommand{\CEGAR}[0]{CEGAR}

\renewcommand{\vec}[1]{\overrightarrow{#1}}

\newcommand{\writtenNumRandProjectVecs}[0]{five}

\newcommand{\numOperators}[0]{101}

\newcommand{\numStateAttributes}[0]{k}

\newcommand{\numberSingleFeatureDistanceSelectors}[0]{54} 

\newcommand{\numberProductSelectors}[0]{1552}

\newcommand{\numberOfSelectors}[0]{1731}

\newcommand{\autoUserVecLength}{k_{au}}

\newcommand{\varForNumberBadRefinementsAutouserIsAllowed}[0]{c_{au}}

\newcommand{\seqNotation}[2]{\langle~#1~\textbf{|}~#2~\rangle}
\newcommand{\ithval}[2]{\{#1\}_{#2}}

\newcommand{\githubForCodeFromThisPaper}[0]{\url{https://github.com/DBay-ani/Operator_Selection_Learning_Extensions_For_Fanoos}}

\begin{document}

\title{A Learning-Based Method for Automatic Operator Selection in the Fanoos XAI System} 

\titlerunning{Learning-Based Operator Selection in Fanoos}

\author{David Bayani}

\authorrunning{David Bayani\orcidID{0000-0001-5811-6792}}
\institute{Computer Science Department\\Carnegie Mellon University, Pittsburgh PA 15213, USA\\
\email{dcbayani@alumni.cmu.edu}
}
\maketitle

\begin{abstract}

We describe an extension of the Fanoos XAI system (\cite{https://doi.org/10.48550/arxiv.2006.12453,DBLP:conf/vmcai/BayaniM22})
which enables the system to learn the appropriate action to take in order to satisfy
a user's request for description to be made more or less abstract. Specifically, descriptions 
of systems under analysis are stored in states, and in order to make a description more or less 
abstract, Fanoos selects an operator from a large library to apply to the state and generate a
new description. Prior work on Fanoos predominately used hand-written methods for operator-selection;
this current work allows Fanoos to leverage experience to learn the best operator to apply in a particular
situation, balancing exploration and exploitation, leveraging expert insights when available, and utilizing similarity
between the current state and past states. Additionally, in order to bootstrap the learning process (i.e., like in
curriculum learning), we describe a simulated user which we implemented; this simulation allows Fanoos to gain
general insights that enable reasonable courses of action, insights which later can be refined by experience with real users, 
as opposed to interacting with humans completely from scratch.
Code implementing the methods described in the paper can be found at \githubForCodeFromThisPaper.

\end{abstract}

\section{Introduction}

Explainable artificial intelligence (XAI) has garnered increasing attention over the last
decade, a surge often attributed to the increased power --- and therefore desire to use ---  AI 
systems based on machine learning (ML), whose decision making processes are typically difficult  
for human users to understand. In many fields of practice, due to ethical, safety, and legal concerns, 
there has been hesitance to adopt the latest ML technology, in no small part due to the inability of
area experts to check that the software is acting in 
an appropriate  manner. 
Inspired by scientific and practical needs, Fanoos is an XAI system developed to produce explanations
at multiple levels of abstract to suit a user's situational needs. Fanoos allows users to interactively ask questions
about an ML system's behavior and receive explanations which, at the user's request, can be made either more or less
abstract. Further still, explanations provided by Fanoos come in multiple strength which users may freely choose 
between: Fanoos can provide descriptions that are guaranteed to reflect the system's true behavior in all circumstances
(including extremely rare pathological situations that may be unuseful to consider in practice), or, at the user's
request, Fanoos can attempt to explain the typical behavior of the system, capturing common-case occurrences without
being bogged-down by pathological cases or circumstances that have zero probability of occurring.

In this work, we detail an extension of Fanoos which allows the system to learn what actions to take in order to
better satisfy a request for greater or lesser abstraction. In particular, explanations shown to users are stored
in \FStates~tracked by Fanoos, and modifying explanations is done by Fanoos selecting then apply an operator on 
an \FState~ in order to generate a new \FState. In this paper, we overview our approach to enabling Fanoos to learn
which operator to apply to a given \FState~in order to satisfy a user request, automatically balancing insights from
previous user interactions, advise from expert-provided heuristics, and the need for sufficient exploration.
Taking inspiration from curriculum learning, we also implement a simulated user which serves to bootstrap 
Fanoos's learning process, allowing time with real humans to be spent fine-tuning what it has learned as opposed
to starting from scratch.

Code for our approach can be found at \githubForCodeFromThisPaper.

\section{Brief Aside: Some Comments on the History of this Paper}

The content of this paper is based on the ideas-document from early 2020 found under UUID
\seqsplit{d7fecc3b-93bb-424e-a838-3f000f3715cf}
at \url{https://github.com/DBay-ani/FanoosFurtherMaterials/blob/master/manifest.xml}.\footnote{
An immutable, time-stamped version is available at 
\url{https://web.archive.org/web/20210717224136/https://github.com/DBay-ani/FanoosFurtherMaterials/blob/master/manifest.xml}
} 
Publicly, 
brief written mention of these
endeavors appeared in
Appendix A.1.1 of \cite{https://doi.org/10.48550/arxiv.2006.12453}
and
A.1 of \cite{IJCAIXAIFanoos2020}; 
I discussed them in greater detail 
during personal interactions
at ICAPS-XAIP 2020 and IJCAI-XAI 2020.\footnote{\url{https://web.archive.org/web/20220508212351/http://xaip.mybluemix.net/2020}}

\section{Overview}
\label{sec:overview}

In this section, we provide an overview of how operators are selected in
Fanoos, as well as the design criteria that governed them. The approach
taken allows for leveraging expert rules for selecting operators as well
as learning which operator to take.  It supports classic logic-based rules,
as well as approaches more closely related to soft classifiers.

The basic idea is as follows: 
\begin{itemize} 
\item{There is an indexed set of operators, $S_O$, which may be applied to a
    \FState.} 
\item{There are an indexed set of selectors, $S_s$, which each produce a 
    (normalized) distribution over $S_O$. Each selectors takes in a variety of
    information, including the entire history of use in Fanoos, and the current 
   \FState~being applied to. 
    For the sake of simplicity, we will write this as 
    $s_i(*)$ where $s_i \in S_s$ --- the point here is that $*$ is used to 
    represent the variety of arguments $s_i$ may have.} 
\item{A process weighs and combines the distributions produced by members of
     $S_s$ to produce a final distribution over $S_O$.} 
\item{The final distribution is used to inform the selection of operator.  
    This selection process is conducted in a way that
    promotes a healthy 
    balance between exploration and exploitation.} 
\item{After operator  application,  \FState~ formation, and receipt of the user's
    next request, Fanoos is internally provided a numeric score as feedback, this score being 
    based on the user's request in the context of the queries preceding it.} 
\item{The numeric feedback is used to adjust the weights given to each member 
    of $S_s$, 
    and, in order to inform our exploration process, other bookkeeping 
    is done to keep track of how often each operator has been used.
    }
\item{This process repeats until the user exits.} 
\end{itemize}

We list now some of the high-level ideas that have shaped this design:

Generally, there are three categories of information that are available which we would like to
leverage: 
\begin{enumerate}[label*=(A\arabic*)] 
\item{ \label{considerationA1} \textit{Number of times each operator was tried}
    - we want to make sure we explore sufficiently} 
\item{ \label{considerationA2} \textit{Success rate} - we want to try and pick
    operators that work well, for our notion of
    what ``working well'' is.}
\item{ \label{considerationA3} \textit{Distance} - In addition to having access to a \FState~which we
   want to apply and operator on, and we have
   all the prior \FStates, operators applied to them, and the results of operator application from the
   past. We would like to leverage knowledge of previous \FState's structure
   to decide what to do, as opposed to simply choose operators based on what
   tends to work well when averaged across all \FStates. As such, 
   knowledge of how far different a \FState~ is from each prior \FState~ can be used
   to help inform 
   the decisions as to  which operator is best applied in the current situation.} 
\end{enumerate}

Ideally, the approach could allow for both of the following to
be worked-in effectively: 
\begin{enumerate}[label*=(B\arabic*)]
\item{\label{considerationB1} Learning of which operator to apply}
\item{\label{considerationB2} Expert knowledge and guidance on which operator to
    use} 
\end{enumerate}

The approach taken addresses each of \ref{considerationA1}-\ref{considerationA3} 
and \ref{considerationB1}, \ref{considerationB2} above.

\section{Operators}
\label{sec:operators}

Fanoos produces descriptions in response to user's questions based on 
values it stores in its most recent \FState. 
Inside \FStates~ are stored \CEGAR-refinement parameters, constraints on which 
predicates are allowable inside descriptions, various settings for parameters that
influence the description generation process, and the content of the \FState's
user-facing description itself.
In order to accommodate a user's request that a description's abstraction level be changed,
Fanoos selects then applies
an \textit{operator} to the most recently used \FState, generating a new
\FState~ from which to base a new description. 
In general, we view the process of responding to a user's requests for changing
abstraction level as being analogous to a binary tree search ---
where nodes are \FStates~ and paths are determined by user's requests. Operators
act as the actual mechanism that moves current attention from a parent node to the child node most
suitable for the corresponding user request, in particular triggering the 
generation of the \FState-description in the process.

We divide the \numOperators~operators in our implementation into three categories: special operators, 
parameter-adjustment operators, and predicate constraining operators.

\subsection{Special Operators}
\label{subsec:specialOperators}
%
In our implementation, we consider two special operators: 
the start operator and the blank operator. The start operator
is used exclusively to generate the initial description following a user's question,
using default settings for all internal parameters;\footnote{In principle, 
one could attempt to tune the starting \FState~ values to maximize the proportion of 
cases where users are satisfied by the first description provided and
do not request any further adjustments. We do not pursue such an extension in this 
work, however.} of the operators we have implemented, the start operator is the 
only one that may fill this role. The blank operator functions by simply re-running
the description generation procedure over the \FState~ without modifying any aspect
of the \FState~except for the fresh generation of all reachability results\footnote{Unless otherwise
specified by parameters of a \FState, Fanoos reuses reachability results that are stored in the previous \FState, both
for increased efficiency and to better control sources of variability in description generation.}
and the associated
description content. Given the non-determinism present in components 
of Fanoos, the blank operator helps measure the natural variance of the system's behavior
and establishes a baseline for how often improvements occur purely due to chance.
Providing insights for testing aside, the blank operator is a reasonable course of 
action for situations where Fanoos has ``essentially correct'' settings in the \FState, 
but the description would benefit from slightly different choices among alternatives that 
prima facie seem equally good.

\subsection{Parameter-Adjusting Operators}
\label{subsec:parameterAdjustingOperators}

Parameter-adjusting operators, as the name entails, modify internal parameters of \FStates,
in turn influencing various aspects of the process that ultimately lead to the description Fanoos
presents to users. The parameter-adjusting operators implemented
modify or set various combinations of the following parameters:
\begin{itemize}
\item{The sampling radius scaling parameter used during the process of determining the subset of predicates
    consistent with a box that are most specific ($\alpha$ in appendix E of \cite{https://doi.org/10.48550/arxiv.2006.12453}). } 
\item{ Whether to reuse previous reachability results as a starting place for the 
    reachability analysis needed by the current \FState, or to freshly compute all
    results.}
\item{ Whether during the \abstractState~ refinement process, \abstractStates~ (i.e., boxes in our
    current implementation ) should have their axes split only along variables that appear
    in the question, or whether all variables should be candidates for splitting, regardless of whether they
    are involved in a user query. This determines whether the variable $h$ from
    appendix F.6.2 of \cite{https://doi.org/10.48550/arxiv.2006.12453} 
   should be allowed to range over all possible values
   or only a particular subset. }
\item{ Whether to attempt merging boxes 
    after the reachability analysis,\footnote{Any merging occurs after the
    reachability analysis but \textit{prior} to trying to fit predicates to boxes.} and if so, how many iterations
    of merging should be tried. In our primary box-merging algorithm, one iteration examines each of the box-corners
    available that have at least two boxes incident on it to see if any of the incident boxes can be merged; multiple
    iterations repeat this process, each time using the updated list of available corners from the previous timestep.
} 
\item{ The degree of precision to use when comparing coordinate values if merging boxes. 
	As explained in section 2.2 of \cite{https://doi.org/10.48550/arxiv.2006.12453}, multiple boxes
	may be merged into a single box of slightly larger net volume, with the amount of permissible expansion
	determined by a precision threshold;
	this process allows us to merge boxes that roughly align (but might not exactly align), while
	preserving the soundness of our guarantees.}
\item{ The side-length used to determine when refinement should stop (that is, 
     $\epsilon$ in equation 2 of \cite{https://doi.org/10.48550/arxiv.2006.12453}).} 
\item{ The value for Boolean variable "produceGreaterAbstraction" in algorithm 3 (the function "generateDescription")
    of \cite{https://doi.org/10.48550/arxiv.2006.12453}. }
\end{itemize}
The complete list of operators and how they modify the \FState~parameters can be found at
the public GitHub repo containing the code (\githubForCodeFromThisPaper).

\subsection{Predicate Constraining Operators}
\label{subsec:predicateConstrainingOperators}

Predicate-constraining operators effect what predicates are allowed to be used in forming descriptions, 
either disallowing certain predicates or re-allowing predicates that were previously barred.
In total, there are four such operators, accounting for all combinations of either allowing / disallowing 
a predicate and whether the aim is to increase or decrease the abstraction level.
We will detail how predicates are selected for removal, then comment on how the process differs when
making  decisions to  re-add them.

Let $\SymbFState_T$ be the \FState~
whose description, $\SymbFSDescr_T$, the user currently wants altered.  Let $\SymbUQ(\SymbFState)$ be the 
specific question instance\footnote{Here, if the same question is asked later, it is considered
a different instance.} 
for which, in the process of producing replies, a \FState~ $\SymbFState$~
was generated. 
To determine which named predicate occurring in $\SymbFSDescr_T$ to remove, 
the records of previous interactions are examined in order to select a candidate that best 
balances exploration with exploitation.\footnote{
Exploration: trying the available options often enough to be informed of each potential outcome;
Exploitation: choosing the option that, based on the information accumulated so far, seems most 
likely to result in the outcome the user requested --- changing the abstraction
level in the desired direction.}
Given a \FState~ that occurred in the past, $\SymbFState_t$,
let:
\begin{itemize}[nosep]
    \item{ $\omega(\SymbFState_t, p)$ be the number of times a named predicate, $p$, occurs in the description of \FState~ $\SymbFState_t$. This may be greater than one if, for instance, p occurs in multiple conjuncts.\footnote{Naturally, one can consider a variant of this where, in what follows, one uses $\mathbbm{1}(\omega(\SymbFState_t, p) > 0)$ instead of $\omega$ raw. Our implementation does not 
work in such a fashion, but one can of course implement such an operator instead of --- or in addition to--- what we currently have
in our code.}}
    \item{$rm(\SymbFState_t)$ and $rl(\SymbFState_t)$  be predicates indicating that the user requested the description to become, respectively, more abstract and less abstract (rm: ``request more'') }
    \item{$rb(\SymbFState_t)$ indicate that the user requested to exit the
    inner QA-loop (i.e., ``b'' in Listing 1.3 of \cite{https://doi.org/10.48550/arxiv.2006.12453})
    after seeing $\SymbFState_t$'s description (rb : ``request break'')}
\end{itemize}
Further, let 
$r_T = rm$ and $r_{T+1} = rl$ if the user requested that $\SymbFSDescr_T$ (the current description)
 become more abstract,
and $r_T = rl, ~r_{T+1} = rm$ if the user requested lower abstraction.
The predicate to remove is determined using the index returned by
\setlength{\abovedisplayskip}{0pt}
\setlength{\belowdisplayskip}{0pt}
\setlength{\abovedisplayshortskip}{0pt}
\setlength{\belowdisplayshortskip}{0pt}
\begin{equation}\label{eq:predicateConstrainingOperatorUCB}
\begin{aligned}
\text{UCB}(& ~\big < |\text{occ}(p)| ~ \big | ~ p~\text{occurs in}~\SymbFSDescr_T \big >,\\
	& ~\big < |\text{succ}(p)| ~\big | ~ p~\text{occurs in}~\SymbFSDescr_T \big > ~ )
\end{aligned}
\end{equation}
where UCB is the deterministic Upper Confidence Bound algorithm \cite{DBLP:conf/focs/Auer00} and 
\begin{align*}
\text{occ}(p) &= \{\SymbFState_t \in \SymbHistory_t | r_{T}(\SymbFState_t) \land (\omega(\SymbFState_t, p) > \omega(\SymbFState_{t+1}, p)) \}\\
\text{succ}(p) &= \{\SymbFState_t \in \text{occurs}(p) | r_{T+1}(\SymbFState_{t+1}) \lor rb(\SymbFState_{t+1}) \}
\end{align*}
where ``$\SymbFState_t \in \SymbHistory_t$'' is a slightly informal reference to accessing $\SymbFState_t$ from \textit{all} previous interaction records
(i.e., not just replies about $\SymbUQ(\SymbFState_t)$ or records from this user session).
$\SymbFState_{t+1}$ indicates the \FState~ that followed $\SymbFState_t$ \textit{while responding to the same question, $\SymbUQ(\SymbFState_t)$} (i.e., it is not simply
any \FState~ that comes chronologically after $\SymbFState_t$ in database records); 
in the cases where $\SymbFState_{t+1}$ does not exist, we substitute infinity for $\omega(\SymbFState_{t+1}, p)$,
and false for both $r_{T+1}(\SymbFState_{t+1})$ and  $rb(\SymbFState_{t+1})$.
Operators that re-allow predicates follow a very similar process as the above, except 
the direction of inequality in the definition of $\text{occ}(p)$ is reversed, and
instead of considering the predicates that \textit{do} occur in $D_{T}$ in \cref{eq:predicateConstrainingOperatorUCB}, only the predicates
that were \textit{forbidden from occurring} in $D_{T}$ are considered.

While alternatives to the adopted method could be used --- for example, approaches with greater
stochasticity --- we believe our choice of a UCB algorithm is 
most likely
appropriate at this stage, considering its relative data efficiency and the 
likely nature of the environment.\footnote{For instance, we do not expect an
adversarial environment.}
Future improvements or novel operators may introduce different or more sophisticated methods for
predicate selection, such as attempts to further leverage joint-relationships present between predicates
in a descriptions and/or contexts.

\section{Inference}
\label{sec:inference}

This section provides details on how operators are selected; that is, how
inference is performed.

Let $w_{i} \in \mathbb{R}$ be the weight that selector $s_i \in S_s$ is
given by the system (this value may be negative), and for any non-negative integer $m$ let
\[ [m] = \{m' \in \mathbb{N} \setminus \{0\} | m' \le m\} \]
We form the following distribution, $D_{samp}(*, \vec{w})$, which we will use momentarily to inform
the selection of operator:\\
\begin{equation}
\label{eq:inference:DSample}
\begin{aligned}
D'_{samp}(*, \vec{w}) = \underset{i \in [|S_s|]}{\sum} w_i s_i(*)\\
d' = \underset{i \in [|S_s|]}{min}(\mathbbm{1}( \ithval{D'_{samp}(*, \vec{w})}{i} \le 0)\ithval{D'_{samp}(*, \vec{w})}{i}) \\
D_{samp}(*, \vec{w}) =
	\frac{D'_{samp}(*, \vec{w}) - d'\vec{1}}{
        \big(\underset{i \in [|S_s|]}{\sum}w_i\big) - d'|S_o|}
\end{aligned}
\end{equation}
The subscript ``samp'' on $D_{samp}$ is short for ``sample'', a name reflective of
roughly how we use it next.
In the above, we only subtract the minimum weight when it is negative ---
so a glut of positive weights can actually
flatten the distribution.
The Upper Confidence Bound (UCB) algorithm (\cite{DBLP:conf/focs/Auer00}), widely used from one-arm bandits,
is then applied to $D_{samp}(*, \vec{w})$ as though each member of
$D_{samp}(*, \vec{w})$
gave a success rate; this algorithm is responsible for producing the index
of the operator to use. Notice that
by using the (UCB) algorithm, we address \ref{considerationA1} while potentially
respecting \ref{considerationA2} and \ref{considerationA3}. That is,
the selectors themselves will address \ref{considerationA2} and
\ref{considerationA3} when forming their vote distributions,
while we address \ref{considerationA1} at the very end by using the UCB
algorithm to balance the selectors'
suggestions with necessary exploration.

Observe that in our inference procedure,
the only way we know anything about the \FState~ is through the votes provided by the selectors.
Another fact worth highlighting is that, from the standpoint of theoretical concerns, 
we violate the assumptions of the UCB algorithm. The UCB algorithm provides guarantees
under the assumption that the world state does not change --- i.e., that the ``success rate''
for each operator, while not directly observed, is constant. We use this basic bandit algorithm
in order to facilitate a simple implementation that explicitly considers the 
factors we highlighted (e.g., \cref{considerationA1}); we have few qualms if one wishes to substitute-in 
a more sophisticated method at this inference phase (e.g., contextual bandit methods), so long as it satisfies our general requirements.
On balance, here we have provided a straight-forward
approach that is reasonable from a mechanical standpoint.

If an operator is ultimately selected that is not applicable --- for example, an operator
that tries to remove named-predicates when the state it is being applied to has a description that lacks any --- then $q_{t+1}$
copies the description and pertinent parameters from $q_{t}$ ( some internal content must
differ between $q_{t+1}$ and $q_{t}$, such as the histories and bookkeeping data used
in our specific implementation).

\subsection{Supporting Further Featurization}
\label{subsec:inference:furtherFeaturization}

Ignoring the details of the normalization done to form it,
$D_{samp}(*, \vec{w})$ can be seen as essentially a product between
the weight vector, $\vec{w}$, and a matrix containing the distribution of votes
for each selector. Thus, on its surface, it would seem our aggregation
scheme is
more-or-less linear, and thus would be unable to leverage joint-behavior
of selectors. While accurate on a shallow assessment, we detail in this subsection 
how we have enabled more sophisticated inferences by properly crafting of the
vote-gathering process and the scope of information visible to selectors.  
Essentially, 
we introduce ``features'' based on a subset of selectors' votes, making a 
scheme that is linear in respect to the feature space but potentially 
non-linear in respect to the original space of selector votes.

In order to gather votes from selectors, each selector is queried to
determine if it is ready to provide its vote-distribution over the
operators. This occurs in a loop: for each iteration of the loop,
at least one selector must cast its vote,
and selectors may not redeliver, modify, or rescind
their votes once they are cast. 
Since 
selectors have privy to a broad base of
information, they may see the vote-distributions issued thus far
by their co-patriots. Using this fact, we can easily support featurization
by specifying a base-set of selectors that produce the ``raw signals'', and
a set of meta-selectors whose votes are purely functions of the votes
provided by the base-set. See, for instance, the implementation of second-order
selectors discussed in \cref{sec:selectorsTried}.

The method used to segment blocks of selectors can be compared to a variety
of techniques from ML and rule-based systems. In \cref{appendix:interpretationsOfOpSelectionProcess}
we describe some interpretations of this approach; each interpretation, while referring to the
same implementation and raw facts, does provide different intuitions, different connections to prior work, and
different insights for further exploration.

\subsection{General Comments and Future Work 
    for Our Vote Aggregation Scheme}
\label{subsec:inference:generalCommentsAndFutureWork}

We are not overly committed to the use of \cref{eq:inference:DSample} for vote
aggregation, and may modify it further in the future. \Cref{eq:inference:DSample}
is largely just a normalized weighted sum, and as such has some 
natural motivations
and interpretations. This said, it is arguable that the form of \cref{eq:inference:DSample} 
does not appropriately match how it is utilized in the UCB algorithm --- for instance, it
provides a sum that is normalized in respect to all operators, when the UCB algorithm is designed
to deal with the individual success rate of each operator.\footnote{Mechanically this is not a show-stopping
issue, and most likely is not problematic from a theoretic standpoint in the limit. However, 
it does open-up more scenarios where the final operator is decided primarily based on exploration
considerations as opposed to exploitation (i.e., when the addition of uncertainty bounds changes the ranking
of operators).}
While there is obviously room for an inference
rule more grounded in theory or for deeper connections (and justification) from existing literature,
at this time we content ourselves with a procedure for this step that is generally sensible (while perhaps
imperfect) and effectively incorporates the categories of information we wish to consider.

An aspect of the approach taken in \cref{eq:inference:DSample} which we would like to retain
is the fact that we essentially get ``reverse selectors'' for free --- that is,  if $s_j$ is a selector whose
pattern of voting is negatively correlated with good actions, 
the system uses votes from $s_j$
to \textit{reduce} the likelihood of doing what $s_j$ suggests. An alternative we may consider is
to use a Winnow-like update (e.g., \cite{DBLP:conf/dagstuhl/Blum96}),
as opposed to the rule adopted in \cref{eq:updateForSelectorWeightAtTimeT}; such a modification would 
keep all weights positive and more comfortably ignore ```irrelevant'' selectors, but
would require the incorporation of additional selectors explicitly providing ``reverse'' votes\footnote{
These ``reverse selectors'' could be implemented as higher-order selectors that take the vote distribution
of one selector, $s_i(*)$, and output 
$$\seqNotation{f(j)\big(\sum_{j \in [|S_o|]} f(j) \big)^{-1}}{j \in [|S_o|]}$$
where 
$$f(j) = U\big(\alpha \ithval{ p_i(*) }{j} + (1- \alpha) U\big)^{-1}$$
for some hyperparameter $\alpha \in (0,1)$.}
in order
to emulate the ability to leverage selectors whose votes tend to negatively correlate with proper actions.
Another option worth bearing in mind that has similarly shaped parts is the update method used in \cite{rabbany2018active}.

In future work, we may further examine the use of boosting-based approaches 
(e.g., soft-classifier variants of AdaBoost \cite{DBLP:conf/eurocolt/FreundS95})
to produce a vote aggregation scheme that results in stronger overall outputs.
In terms of engineering improvements,
our current vanilla implementation of the selectors-voting framework
is entirely serial in execution, but the process is trivially
parallelizable,\footnote{Up to the explicit dependency of higher-order selectors on 
lower-order selectors to vote first, of course. If we view these dependencies between
selectors as a DAG, then each layer can run in parallel. For our implementation, 
and most reasonable extensions we can foresee, the width of layers (which is parallelizable
work) is far larger than the number of layers (serial bottleneck). In combination with parallelization,
probabilistically ignoring selectors based on their weight may also be an option if the run-time of this
part of the procedure ever becomes a concern.} a fact that ideally
would be leveraged.

\section{Learning}
\label{sec:learning}

We will break our discussion of learning from feedback into two parts: how
the selector-weights are updated given a reward signal, and how the reward
signal is derived from user feedback.

\subsection{Updating Selector Weights Given the Reward Signal}
\label{subsec:learning:updatingSelectorWeights}

Suppose we are given a reward signal $y_t \in \{-1, 1, 0\}$ at iteration
$t \in \mathbb{N} \setminus \{0\}$ (in the next section, this signal is
derived from the user feedback and the history, but for now we suppose a
numerical score is already derived).  Let $O_t$ be the operator that was actually
selected for iteration t (i.e., the one chosen at the end of inference).
Tweaking notation slightly to incorporate a time-index, we update the weight
of each selector as follows: 
\begin{equation}
    \label{eq:updateForSelectorWeightAtTimeT}
    w_{i,t+1} = w_{i,t} + y_t(~~(s_i(*_t))(O_t) - U)
\end{equation}
here, $U$ is the
probability that $O_t$ would have been chosen at random ( given a uniform
distribution over $S_o$), and is simply $\frac{1}{|S_o|}$. 
Basic sanity checks of \cref{eq:updateForSelectorWeightAtTimeT} reveal the 
sensibility of subtracting $U$.
To start, it aligns with the intuition that we
positively (respectively, negatively)
 reward selectors that gambled more than random on an operator that ended
 up being correct
(respectively, incorrect). By similar reasoning, this update rule deals sensibly with  
selectors that ``do not commit'' in favor or against the selected operator  
by neither directly promoting nor directly demoting such selectors\footnote{We use ``directly'' in reference to the 
fact that, relative to the updates given to other selectors and given how votes are aggregated 
(\cref{eq:inference:DSample}), a selector may loose or gain sway over which operator is ultimately selected even if its
individual weight is not modified.}; 
this has noteworthy practical benefits, since it supports
selectors that act as ``area experts'' --- having strong and well-informed opinions
for some circumstances, but being allowed to abstain without \textit{risking} punishment
when they believe a situation is outside their purview.
Further still, it 
assists in reducing the impacts of any "coasting" on the success
of other selectors: when a selector that did not ``help cause'' the selection of a correct (incorrect) operator 
receives a reward (punishment) in the form of increased (decreased) weight solely as a result of a 
successful (unsuccessful) operator 
having ultimately been 
used at the end of that iteration.\footnote{Consider how a selector that 
consistently votes the uniform distribution would be treated were subtraction of $U$ absent, such as when the system as a whole is
more often successful than not. Bear in mind that while
the addition of a uniform value would not alter the ranking of operators under the aggregated vote (
\cref{eq:inference:DSample}), the contribution would serve to "flatten"  
$D_{samp}$.}

As a closing note for this subsection, we mention in passing a 
possible connection this update rule has to ( and hence lessons-learned that can be leveraged from )
variance reduction via baselines in vanilla policy gradients (e.g.,
\cite{Williams92simplestatistical}). 
For instance, (1) the "shape" and, in
some sense, "content" of our update are not too dissimilar in appearance
from the vanilla policy gradient, (2) we clearly have a baseline term
($U$), and (3) like in PG, 
the action we take is not solely determined by arg-max of our final distribution, 
but incorporates considerations for exploration.\footnote{PG randomly samples the final distribution, whereas we use
UCB to more explicitly account at this final step for the number of times an operator has been tried.}
The approach proposed in this write-up was constructed prior
to making this sort of connection, but it is no surprise that similar proposals
might appear in similar problem settings --- however, more work is to be done
in order to certify that there is more than a superficial connection, one that offers more 
insight than citing the general relevance of the entire sub-field of model-free RL.

Further modifications to \cref{eq:updateForSelectorWeightAtTimeT} may come by more fully 
approximating an error gradient based off of \cref{eq:inference:DSample}, which would require 
\cref{eq:updateForSelectorWeightAtTimeT}  to have a multiplier of roughly 
 $\frac{\sum_{j, j\not=i}w_{j,t}}{(\sum_{j}w_{j,t})^2}$ for most $i \in [|S_s|]$ (ignoring special effects of negative terms);
in addition to the standard mathematics that would motivate this, it also plays into the intuitive story
of punishing or rewarding selectors based on ``how much they caused an outcome'', where a selector's weight 
certainly plays a role in its blame. 
Space for modifications and alternatives abound, depending on how much of the rest of the system one is willing to 
modify --- some such options we discussed in \cref{subsec:inference:generalCommentsAndFutureWork}. Here, as we did there,
we consider an effective approach that meets our general criteria, but which we do not claim is optimal or beyond criticism.

\subsection{Producing the Reward Signal}
\label{subsec:learningProducingTheRewardSignal}

We now discuss how, in this initial implementation, we form the $y_t$
used in the previous subsection. Below, for ease of discussion, we will use
"iteration t" and "time t" interchangeably.

At a high-level, the process in Fanoos follows the following overall flow: 
\begin{enumerate}[{label=(\arabic*)}]
\item{\label{subsec:learningSteps:1:makeState} Produce the \FState~ for time t.} 
\item{\label{subsec:learningSteps:2:showResponce} Display a response for time t by
    extracting relevant information from iteration-t's \FState.}
\item{\label{subsec:learningSteps:3:getFeeback} Receive user-feedback for time t, $f_t$.}
\item{\label{subsec:learningSteps:4:chooseOperator} Choose an operator at 
    iteration t to be used to form the \FState~for time $t+1$.}
\end{enumerate} 
The types of user feedback currently supported are listed in
\cref{tab:typesOfUserFeedback}. 
Once
we apply the operator to produce a new \FState~and show it to the user, we will
receive more user feedback, $f_{t+1}$. While many different methods for forming
$y_t$ can be supported, our implementation uses the reward function
described in \cref{tab:rewardFunction}. The idea behind this reward
function is that we want to learn which operators do in fact produce more
(respectively, less) abstract descriptions on user demand. We take the user
"reversing" their request (e.g. $f_t=l$ and $f_{t+1}= m$) to be a sign that
abstraction definitely went in the correct direction and to a non-trivial
degree. Similarly, if $f_{t+1} = b$, we take it as a sign that we satisfied
the user's previous request.\footnote{\label{footnote:futurework:exitButIndicateDisatisfied}
For those interested, it would be trivial to extend the user interface with an option, perhaps under the 
submenu invoked by entering ``u'', to exit the description adjustment loop but interpret the exiting as
failure. Ideally, the user would not have to be concerned about such things and any confusion
caused to Fanoos by \textit{not} having this additional option would be minimal.}
Naturally, the simplicity of this approach comes at
the cost of potentially missing more nuanced patterns,
such as what the appropriate feedback should be in hypothetical cases where we "got close to the right
abstraction level but overshot/undershot". At the very least, however, our proposed reward function appears
to be a reasonable way of guiding operator selection for cases that are less murky.

\begin{table*}[h]
        \begin{tabularx}{\textwidth}{p{1cm}|p{\textwidth - 1cm}}
        \hline
		\multicolumn{2}{|c|}{Types of User Feedback} \\ \hline
short-hand & description \\ \hline
l & make the description less abstract \\ \hline
m & make the description more abstract \\ \hline
b & ends the user's interrogation regarding the current question. \\ \hline
u & allows the user to specify operators to apply or, in the case of the history-travel or manual predicate review operators, interact with. \\ \hline
\end{tabularx}
\caption{Description of the types of user responses Fanoos currently supports}
        \label{tab:typesOfUserFeedback}
\end{table*}

\begin{table*}[h]
\begin{center}
        \begin{tabular}{|c||c|c|c|c|c|c|c|c|c|}
        \hline
                \multicolumn{10}{|c|}{Reward Function} \\ \hline
		\textbf{$f_t$}  & m  & m  & m  & l  & l  & l  & m/l  & u  & b   \\ \hline 
		\textbf{$f_{t+1}$}  &  m  &  l  &  b  &  l  &  m  &  b  &  u  &  (any)  &  (none)  \\ \hline 
		\textbf{$y_t$}  &  -1  &  1  &  1  &  -1  &  1  &  1  &  0  &  0  &  (none)   \\ \hline 
\end{tabular}
\end{center}
\caption{Description of the reward function Fanoos currently uses, showing the reward signal as a 
function of consecutive user requests. Note that when the user enters "b" at iteration t, the 
session of question-and-answering breaks, so there is not feedback or reward signal from iteration $t+1$, since 
there is no iteration $t+1$.}
\label{tab:rewardFunction}
\end{table*}

\section{Bootstrapping the Learning Process: The \Autouser}
\label{sec:autouser}

\subsection{Purpose}

Prior to spending the time, capital and patience needed for a human to train the operator
selection system from scratch, we bootstrap the learning process by defining
an oracle 
to act approximately as we expect a human user would; one can view this proposal
as a form of curriculum learning (\cite{https://doi.org/10.48550/arxiv.2003.04960}), 
where the \autouserFull~ (``\autouser''~or~``\autouserShort'') provides the first set of tasks (and hence lessons)
to \methNam~ so that \methNam~ can more easily and rapidly fine-tune itself to satisfy real
users' requests.
This
proposal bares a few points worth clarifying:

First, the fact that we define an "\autouser" to
     evaluate responses of \methNam~ does \textit{not} itself entail
     that we could hard-code an operation-selection method that optimizes
     it directly. The fact the one knows how to evaluate a result does not mean
     they know how to produce a result --- this distinction, for instance,
     is clearly visible in NP-complete problems. To some, this point may
    seem obvious (indeed, it is at the heart of much of reinforcement learning). However, we highlight it since, all to often, the first
     problem that comes to mind when brainstorming ``how to deliver the right
     thing to a user'' is trying to properly infer what a user wants; our
     point here is that, even if we had a perfect user model detailing every
     pertinent aspect of human cognition, determining
    what steps to take in order to satisfy that request would still,
     in general, need to be solved and can be far from trivial.

Second, the notion of ``bootstrapping'' a learner on a reasonable
    but imperfect surrogate problem (here, trying to please the 
    \autouser) is far from 
    silly. Unlike a human, an \autouser~ is cheap, perfectly repeatable,
    and entirely transparent in regards to its evaluation method. While we
    doubt that an \autouser~ we implement would  capture all the trends in
    human desires with extreme accuracy, we expect it to more often than not
    agree with humans, particularly in cases where a human user would have
    strong sentiments regarding the comparative quality of results. 
    This pattern of 
    pre-training a learner on a closely related, easily accessible set of
    circumstances is core to curriculum learning and many sim-to-real works in
    robotics.
    Better models would provide us better confidence, but reasonable
    models can be useful even if imperfect, and we do not need to perfectly
    replicate a human mind in software in order to yield worthwhile, positive
    outcomes by taking this approach.\footnote{This should be especially apparent
    to any who believe that abstraction level is objective and/or human-independent
    to any degree.}

\subsection{Method}
\label{subsec:autouser:method}

In what follows, we may refer to the \autouser~we implemented as \autouserShort. Let $\SymbFState_t$ be that \FState~
shown to \autouserShort~  at time $t$, and $\SymbHistory_t$ be the full history of interactions
Fanoos has had up to and including
 time $t$, whether that be with the \autouserShort~ or
actual humans. $\SymbHistory$~ contains the full records of all \FStates, all requests, and any
other pertinent side-information involved in prior interactions up to and include time $t$
--- in short,  
a complete transcription of all past information potentially relevant. \autouserShort~ determines 
what feedback it should provide to Fanoos at time $t+1$ (i.e., $f_{t+1}$) based on  $f_t$.
Let $\vec{\psi}(\SymbFState_{t-1}, \SymbFState_{t}, \SymbHistory_{t})$ be a
$\autoUserVecLength$-length collection of functions, where 
$\ithval{ \vec{\psi}(\SymbFState_{t-1}, \SymbFState_{t}, \SymbHistory_{t}) }{j}$  is a function
that produces a summary statistic about the change in Fanoos's reply from
$\SymbFState_{t-1}$ to $\SymbFState_{t}$, which may be evaluated relative to other changes seen over the
course of $\SymbHistory_{t}$.  Essentially, $\vec{\psi}(\SymbFState_{t-1}, \SymbFState_{t}, \SymbHistory_{t})$ is the list of
criteria \autouserShort~ uses to judge whether responses from Fanoos satisfied its request,
and each component --- whose computation we will detail next --- indicates
whether a criterion clearly changed in a certain  direction (values 1 or -1),
or if  the direction of change is not sufficiently pronounced to warrant labeling it one way versus the other
(valued as 0).

Define $\delta.v_j(\SymbHistory_{t})$ as 
$ \{ \SymbFState_{l}.v_j - \SymbFState_{l-1}.v_j | \SymbFState_{l}, \SymbFState_{l-1} \in \SymbHistory_{t}\}$, 
and let $\text{ECDF}(a, A)$ be the empirical cumulative distribution function at a value $a$ base on the finite-sized sample
$A$ --- i.e., $\frac{|\{ b \in A | b \le a\}|}{|A|}$.\footnote{In our implementation, we use reservoir sampling
to compute the ECDF, in principle providing efficiency benefits. This is perhaps premature optimization on 
our part, since computing the ECDF has not yet been a source of excessive slowdowns. In our defense, it seemed something
prudent to get out of the way now and avoid any worries later.} 
In our implementation,
$\vec{\psi}$ is computed as: 

\begin{equation}
\begin{aligned}
\ithval{ \vec{\psi}(&\SymbFState_{t-1}, \SymbFState_{t}, \SymbHistory_{t}) }{j} =   \\
        & \textbf{step}_{0.5,0.5}( \mathbbm{1}(f_t = ``m")) \times  \\
        & \textbf{step}_{0.5,0.5}( \mathbbm{1}(j \not\in \gamma_1)) \times \\
        \Big(~~~\mathbbm{1}(j &\in \gamma_2) \times \textbf{step}_{0.4,0.6}(\\
        & ECDF(\SymbFState_{t-1}.v_{\pi(j)} - \SymbFState_{t}.v_{\pi(j)}, \delta.v_{\pi(j)}(\SymbHistory_{t}) ) ~~~) +  \\
        & \mathbbm{1}(j \not\in \gamma_2) \times \textbf{step}_{0,0}(\SymbFState_{t-1}.v_{\pi(j)} -  \SymbFState_{t}.v_{\pi(j)}) \Big) 
\end{aligned}
\label{eq:vectorPsi}
\end{equation}
where 
\[
\textbf{step}_{a,b}(x) = 
    \begin{cases} 
    1  & x > b \\
    0  & x \in [a,b] \\
    -1 & x <  a
    \end{cases}\\
\]
In the above, $\pi$ is an injective map from $[\autoUserVecLength]$ to a subset of the
fields present in \FStates~ 
that \autouserShort~ bases its decisions on,  
and both $\gamma_1$ and $\gamma_2$ are subsets of 
$[\autoUserVecLength]$ used to gate the behavior of $\vec{\psi}$. 
Notice that \autouserShort~ only accesses \textit{the difference} 
between the new \FState~and the prior \FState's values
as opposed to the values themselves.
%

In our current implementation, the \autouser~considers the following criteria (i.e., the range of $\pi$):
\begin{enumerate}[{label={j=}\arabic*}]
    \item{\label{enum:autouserCriteria:volNamedPredicates} The total volume (after normalization) of 
	    abstract states covered by user-defined atomic predicates (a.k.a., ``named'' predicates)}
    \item{\label{enum:autouserCriteria:volBoxRangedPredicate} The total volume (after normalization)
        covered by box-range predicates}
    \item{\label{enum:autouserCriteria:numNamedPreds} The number of unique named
        predicates used in the description}
    \item{\label{enum:autouserCriteria:numConjuncts} The number of conjunctions
        that appear in the description}
\item{\label{enum:autouserCriteria:numBoxRange} The number of box-range predicates
        used}
\end{enumerate}
The volume information used by 
\cref{enum:autouserCriteria:volNamedPredicates} and \cref{enum:autouserCriteria:volBoxRangedPredicate}
comes from the first component
(``$csToTV_2$'') returned by algorithm 9 (``getVolumesCoveredInformation'') in 
\cite{https://doi.org/10.48550/arxiv.2006.12453}; as highlighted by our verbiage in the above bullets, the volumes used
are normalized by the total volume covered by all the abstract states, 
thus bounding their ranges and \textit{aiding} in their interpretation
independent of the specific domain and question-type in use. With this mapping
of $j$ to specific attributes, we use $\gamma_1 = \{2,3,4,5\}$ and $\gamma_2 = \{1,2,3,4\}$.

Among our experiments, we intend to conduct ablation studies,  making sure to include
analysis where the fields available to the selectors are disjoint
from those  
allowed to the \autouser. 
Various schemes for allocating information to the selectors (and operators) versus the \autouser~ provide worthwhile insights
--- arrangements such as having both 
be able to see the same fields, allowing one to see a subset of what the other has, 
or insisting that their sources are (at least in name\footnote{That is, since the statistics
in question reflect parts of the same state and system, there may be a ``spiritual sense''
where the information is ``not disjoint''--- our concern is ensuring that any connection between 
the fields is ``meaningful'' and a correlation inherit in the properties of the system, not a trivial
connection equivalent to label-leakage.}) disjoint. One must be careful in all scenarios, of course,
to hedge interpretation of performance by consideration of how much ``label leakage'' there might 
have been (e.g., directly optimizing the \autouser
). Note, however, that even if Fanoos had direct access to the criteria and
objective function in the \autouser, it is still \textit{not necessarily} a trivial task
to determine the courses of action needed to produce the desired outcome. As to this latter point, 
once again experiment can shed light on the strength of that barrier (i.e., the degree of  difficulty faced even
when complete information and a perfect user model are available).

Thus far in \autouserShort's process, it has (1) compiled a list of criteria to base its judgment and (2)
indicated if each criterion increased, decreased, or did not
change to a noteworthy degree. Next, \autouserShort~ must boil-down this information into a determination
as to whether the overall change in description abstraction level was in
the proper direction; in the case that the response failed to move in the proper direction, \autouserShort~ must re-issue
the command in order to allow Fanoos to try again. 
\cref{algo:autoUserDeterminesFeedbackToGive} shows how
\autouserShort~ determines what response to issue.
In the algorithm, $S_1$ gives an idea of the prevailing direction of noticeable changes (qualitatively: positive, 
negative, or zero), while $S_2$ indicates the number of aspects
that have noticeably changed. Notice that if $S_2 > 0$ and $S_1 = S_2$ ---
that there were substantial changes to the description provided and all the noticeable changes were
in the proper direction --- 
\autouserShort~ considers its request from time $t-1$ (i.e., $r_{t-1}$) to be sufficiently well satisfied, and
thus for the next timestep ( timestep $t$ ) \autouserShort~ makes the opposite request.  
If, however, $S_2 = 0$ or $S_1 \le 0$ --- respectively, that there were either no substantial changes or the majority of
non-trivial changes were in the wrong direction --- then \autouserShort~ does not 
consider its most recent request to have be satisfactorily addressed, and thus reissues it
so that Fanoos can make another attempt.

The only remaining cases to consider for \autouserShort~ are
when $S_2 > 0$, but $0 < S_1 < S_2$, corresponding to when there are noticeable changes in the
description that are majority positive but not unanimously positive; 
we will refer to this as the \debateableSit~ case.  Taking 
general inspiration from
curriculum learning, \autouserShort~ judges success in this scenario with increasing harshness as Fanoos
demonstrates increased ability to modify descriptions in the fashion requested, and conversely
is more lenient when Fanoos is having difficulty meeting requests. One could say that, as
Fanoos's performance improves, the \autouserShort's ``expectations'' increase, raising the bar for
what is considered acceptably good. Adopting this adaptive criteria
helps provide signals to Fanoos which highlight actions with effects that, while beneficial,
would lead to  insufficiently compelling changes if taken alone. At the same time, the approach applies long-term pressure
in hopes that Fanoos can eventually uncover how to generate fully satisfying alterations to descriptions ---
alterations that may be extremely difficult to stumble upon if  
absolute perfection was demanded
from the beginning.

In our implementation, \autouserShort~ handles the \debateableSit~case by randomly choosing whether to repeat
its previous request or request something different,\footnote{Respectively, 
indicating that the \autouserShort~ is (a) unsatisfied and hence Fanoos failed or that (b) Fanoos succeeded.} with odds increasingly
favoring the former as the degree of success increases. This randomization is natural in the sense that, for humans,
decisions for which evidence does not provide a clear, dominating answer often are influence by momentary mood
and other arbitrary factors, inducing some randomness on the outcome. Mathematically, this randomization allows
\autouserShort~to indicate the relative effectiveness of each  strategy attempted by Fanoos via the
long-term proportion of success a strategy incurs --- this despite the limited set of individual requests
\autouserShort~ can make. Further, randomness in the user requests help the system as a whole explore 
more widely and reduces the likelihood of ``getting stuck'' during an encounter
with pathological situations (
such as cases where a particular request cannot in principle be satisfied or certain types of
hypothetical, undesirable cycles in interactions occur\footnote{
See \cref{appendix:ourJudgementOfCycles} for further comments on cycles and our perspective on them if they occur in our system.}).

As the success rate increases, the bounds on the acceptable $S_1$-to-$S_2$ ratio increase and become more narrow, 
placing higher demands on \autouserShort~ and providing less leniency. In \cref{algo:autoUserDeterminesFeedbackToGive}, in order to  
determine the rate at which the range of cutoffs narrow (approaching a deterministic cutoff), we introduce
a parameter $\ell$. We tune $\ell$ so that at a  global success rate of 60\% (i.e., the success rate taken across all of $H_t$, not
just across that session), Fanoos must produce a description that achieves higher than the minimum possible ``improvement'' ---
that is, higher than the lowest possible positive value of $\frac{S_1}{S_2}$, which is no less than $\frac{1}{\autoUserVecLength}$.
From this, trivial calculation gives that $\ell \le -\frac{\log(\autoUserVecLength)}{\log(0.6)}$.
In particular, we use:
\begin{equation}\label{eq:howEllIsSet}
\ell = -\frac{\log(\autoUserVecLength -1)}{\log(0.6)}
\end{equation}
We highlight that $\ell$ influences only the
lower bound of the judgment threshold; since $\alpha$ is chosen uniformly at random over $[0,1]$
in \cref{algo:autoUserDeterminesFeedbackToGive}, at a 60\% success rate, the expected threshold is:
$$0.3 + 2^{-1}(\autoUserVecLength -1)^{-1}$$
At this expected threshold, there would need to be roughly $\frac{1.3}{0.7}$ as many changes to the description
in the proper direction as there are changes in the wrong direction --- that is, the value of
$\frac{\{i \in [\autoUserVecLength]~|~\ithval{P}{i} > 0\}}{\{i \in [\autoUserVecLength]~|~\ithval{P}{i} < 0\}}$
(where $P$ is from \cref{line:PIsDefined}) must (in expectation) be roughly at least
$\frac{1.3}{0.7}$ in order for \autouserShort~ to consider the description to have changed in the desired ways.
In simplest terms, a success would need at least roughly twice as many positive changes as negative.

\AutouserShort~ continues to request changes in description\footnote{Assuming it is possible to provide an initial description. If
the initial question is impossible to satisfy (i.e., the question regards a circumstance that could never occur), 
then Fanoos ---as is appropriate--- indicates as much and exits.} until one of three conditions are met: 
 (1) the number of user requested adjustments reaches a pre-specified
maximum (determined by a human-set parameter, e.g., 200 user adjustment requests), (2) the description has not
changed for more than a human-set number of consecutive requests, $\varForNumberBadRefinementsAutouserIsAllowed$, or (3)
the description contains only box-range predicates for $\varForNumberBadRefinementsAutouserIsAllowed$ many 
consecutive adjustments. As per our reward function described in \cref{tab:rewardFunction}, when \autouserShort~
issues the exit command, Fanoos will interpret this as a success signal, despite cases (2) and (3) more clearly being
failures; as is the case with a real user, Fanoos is not informed as to what motivates \autouserShort's exit. 
Since occasions of concern should be rare when using the \autouser--- not only because of natural circumstances 
but also due to $\varForNumberBadRefinementsAutouserIsAllowed$ being sufficiently high ---
they should not be overly influential; empirical demonstration of such may be prudent.
We are not overly concerned regarding these rare cases (and \cref{footnote:futurework:exitButIndicateDisatisfied}
suggests next steps were we ever to be).

\begin{algorithm}
\SetKwData{Left}{left}\SetKwData{This}{this}\SetKwData{Up}{up}
\SetKwFunction{Union}{Union}\SetKwFunction{FindCompress}{FindCompress}
\SetKwInOut{Input}{input}\SetKwInOut{Output}{output}

\Input{$\vec{\psi}$ as defined in \cref{eq:vectorPsi}; the history up to this time step, $\SymbHistory_{t}$; 
    the current \FState, $\SymbFState_{t}$; the previous \FState~ $\SymbFState_{t-1}$; \autouserShort's previous request to Fanoos, $r_{t-1}$;
    $\ell$, the value tuned in \cref{eq:howEllIsSet} }  
\Output{A new request from \autouserShort, $r_{t}$, in response to the description Fanoos provided for the most recent \FState, $\SymbFState_{t}$. }
\BlankLine
$P \leftarrow \vec{\psi}(\SymbFState_{t-1}, \SymbFState_{t}, \SymbHistory_{t})$ \label[pseudocodeLine]{line:PIsDefined}\;
$S_1 \leftarrow \sum_{j=1}^{\autoUserVecLength}\ithval{P}{j}$\;
$S_2 \leftarrow \sum_{j=1}^{\autoUserVecLength}| \ithval{P}{j} |$\;
\If{$S_2 == 0$}{
    \tcc{No relevant aspect of the description substantially changed after \autouserShort's most recent request.}
    return $r_{t-1}$\;
}
$\alpha \sampleFrom \text{uniform}([0,1])$\;
$g \leftarrow \text{get\_global\_success\_rate}(\SymbHistory_{t})$\;

\If{$\frac{S_1}{S_2} \ge \alpha g + (1 - \alpha)g^{\ell}$}{ \label[pseudocodeLine]{magicLine}
return $\text{opposite}(r_{t-1})$ 
}
return $r_{t-1}$\;
\caption{Pseudocode for the process implemented by the \autouser~ (\autouserShort) to judge responses from Fanoos.}
\label[algorithm]{algo:autoUserDeterminesFeedbackToGive} 
\end{algorithm}

\section{Selectors Used In Current Implementation}
\label{sec:selectorsTried}

Having described how selectors are used in \cref{sec:inference}, we now describe further the
selectors we have implemented.

Recall that selectors are the components in our process that can incorporate expert knowledge 
(\ref{considerationB1} ), knowledge of how useful operators are 
(\ref{considerationA2}), and how much the current \FState~looks like previous 
\FStates~ (\ref{considerationA3}). 
The output distributions that selectors provide are expressive, in that they
can represent classical logic-based rules and more --- the former by providing verdicts that 
only have support on a subset of the operators.

In our implementation, we consider four categories of selectors: uninformed
selectors, applicability selectors, history-informed selectors, and second-order 
selectors.

\subsection{Uninformed and Applicability Selectors}

\label{subsec:uninformedAndApplicabilitySelectors}

Uninformed selectors are those that do not consider any aspect of the \FState,
either directly or indirectly. In order to provide a baseline, help interpret
our later results,  
and further encourage 
exploration, we provide a selector that simply places a uniform vote on each
operator (i.e., places equal weight on all operators in all circumstances).
Additionally, for the same reasons, for each operator we create a unique corresponding
selector whose sole purpose is to vote exclusively for its assigned operator
(i.e., the selector outputs a distribution with support only on its assigned operator).

Applicability operators produce votes that only indicate when a
specific subset of operators are unexpected to be applicable in a given 
situation. As such, they assist in regard to consideration \ref{considerationB2}.
For instance, consider the set of operators, $S_{O,1}$,\footnote{Subscripts here are used primarily
as distinct labels as opposed to having deeper semantic meaning.} that function by
re-allowing a named predicate that had earlier been disallowed from appearing in
descriptions; if all named predicates are allowable at time $t$, then no member 
of $S_{O,1}$ is applicable at time $t$. In the circumstances where members of
$S_{O,1}$ are unapplicable, then a specific applicability selector, $op_{a,1}$,
would put all its support uniformly over $S_{O} \setminus S_{O,1}$.
In the case where the members of $S_{O,1}$ are applicable, $op_{a,1}$ puts
a uniform vote on all operators. In short, when an operator is unapplicable, it
is clearly undesirable to use them and an applicability operator
indicates such; when an operator is applicable, there might still be better
alternatives in the specific circumstances, and as such an applicability 
operator then makes no claim as to which option is better.
In our implementation, we provide applicability selectors that indicate
when each of the operators in \cref{subsec:predicateConstrainingOperators} would be sensible to use.

While applicability selectors are useful in their own right --- potentially
helping tip the balance in favor or disfavor of
certain operators via their direct, additive influence in \cref{eq:inference:DSample}
--- they are particularly
useful when used in higher-order selectors (\cref{subsec:higherOrderSelectors}), which
combine together the insights of multiple selectors in non-linear fashions.

\subsection{History-Informed Selectors}

History-informed selectors dig deeper into the values stored in a \FState~than applicability selectors, 
casting votes based on a \FState's reachability analysis
results, description content, and historical similarities. Each of the selectors in this
category  learn --- via their own methods --- which operator is best to apply
in a given circumstance, and thus these selectors are yet another component of our
system that fill desire \ref{considerationB1}.

For an initial and reasonable implementation, we consider history-informed  selectors
that proceed via the following steps:  
\begin{enumerate}[{label=\tiny{Step} \arabic*}] 
\item{\label{enum:howSelectorsVote:1:computeDist} Compute a distance between the current \FState~and a relevant subset
	of those seen in the past,}
\item{\label{enum:howSelectorsVote:2:rankFromDist} Use the distance from \ref{enum:howSelectorsVote:1:computeDist}
    to rank \FStates~ (e.g., the  \FState~from the past
    closest to the current \FState, the past \FState~second-closest, etc.),}
\item{\label{enum:howSelectorsVote:3:voteMassFromRank} Determine the mass to give each operator based on the
    operator's success rate weighed by how close (in ranking) \FStates~ it previously operated
    on are to the current \FState.}
\end{enumerate}
First we will overview what schemes we implemented for determining the distance-based ranking
between \FStates~ (\ref{enum:howSelectorsVote:1:computeDist} and \ref{enum:howSelectorsVote:2:rankFromDist}), then we 
will overview how the resulting ranking is used to divy out proportions of a vote across each operator 
(\ref{enum:howSelectorsVote:3:voteMassFromRank}). Prior to this, however, we take a moment to overview
what information is available to base distances on.

\subsubsection{\CapFState~Fields Used in  Distance Calculations}
\label{subsubsec:firstOrderSelector:determiningDistance}
Let $v_j$ be the value of field $j$ in the \FState; that is, some
observable aspect of the \FState~we know about. 
The set of $v_j$ considered by history-informed selectors include: 
    \begin{itemize} 
    \item{ The minimum, maximum, mean, median, standard deviation, total,
    first quartile and third quartile of
    the scaled input boxes' volumes. The scaled input boxes are those whose
    axis-lengths have been divided by the respective axis of the universal
    bounding box, allowing the measures to have increased independence from 
    the specific domain used at a given time.  
    }
    \item{ We use the same summary statics as the previous bullet, except
    over the set:
    $$\{\text{sum\_side\_length}(b) |  b \in \text{scaled\_input\_boxes} \}$$
    }
    \item{The logarithm of the total number of input boxes divided by the maximum possible
        number of boxes generally possible (e.g., not considering the question asked) 
	in our refinement scheme given the refinement parameters
	used and given the domain information of the model being analyzed. Specifically:
        $$log_3(\text{total\_number\_boxes}) + \zeta_{I}\lfloor log_3(\epsilon)\rfloor$$
        where $\zeta_{I}$ is here taken to be the dimension of the input space. We provide a derivation
        of this in \cref{appendix:derivationOfHowNumberOfBoxesAreUsed}.
    }
    \item{Information on the named predicates that occur in a description,
    namely the number of unique named predicates as well as the total 
    number of times named predicates occur (recall that the
    same named predicate may appear more than once in a description if
    it occurs within multiple different conjuncts).
    }
    \item{The number of conjunctions, disjuncts and box-range predicates.
    }
    \item{Both the total volume and unique volume covered by each of the named
    predicates, box-range predicates, and conjuncts, as based on the results from
    algorithm 9 in \cite{https://doi.org/10.48550/arxiv.2006.12453}. As commented on
    in \cref{subsec:autouser:method}, the individual values that make up these sums
    are normalized. 
    }
    \end{itemize}
The fields used above are applicable across all question types and
domains. As such, using these fields allow our
selectors to leverage all prior experiences --- in a not entirely trivial sense, transferring what it learns 
between question types and domains.
This said, it is not a general requirement of our approach that such broadly applicable
fields be used --- the information provided to selectors can be tailored for specific question types or domains
--- we simply choose the broadest net to cast in our current implementation, and save further specialization
for some other venture.

While we currently do not inform selectors as to  the question type
or domain, future work could examine the proper methods of providing such information,
balancing transfer of experiences between qualitatively different circumstances with 
insights provided by the specific setting.

\subsubsection{Computing Distances}
\label{subsubsec:firstOrderSelector:rankingAndVoting}

Let $Q_{f_{t}}$ be the set of \FStates~ recorded in the history such that for
each $q \in Q_{f_{t}}$, the feedback from the user
prior to forming $q$ is the same as the user's current request; e.g., if the
user requested ``m'' now, then
$Q_{f_{t}}$ would contain \FStates, $q_t$, that were previously presented to
users at a time $t$ following the user's ``m'' request at time $t-1$.
We order the members of $Q_{f_{t}}$
by descending distance from the current \FState, based on $\seqNotation{v_j}{j \in [\numStateAttributes] }$ (i.e., the
distance between \FState~A and \FState~ B is base on
$d(\seqNotation{ A.v_j}{j \in [\numStateAttributes] }, \seqNotation{B.v_j}{j \in [\numStateAttributes] })$,
for some metric $d$). 
Let $\textbf{rank}(q_t)$ be the rank assigned to a \FState~$q_t \in Q_{f_{t}}$
by this ordering.

Currently, the filtered set of states that we use to inform decisions --- here, $Q_{f_{t}}$ ---
neither incorporates (i.e., filters by ) the question type nor by information about the specific domain. While this comes with the
benefit of having a larger volume of experience to draw upon from a larger net of circumstances (arguably
a simple form of ``transfer learning'' between situations), it comes at the cost of admitting ``less precise''
information that could ``blur the view''. Future work may consider this further information (the question type and
specific domain), likely in addition to --- as opposed to a replacement of ---  $Q_{f_{t}}$ (e.g., as another set of selectors to use,
or as terms influencing the distance measures).

Among our selectors, \numberSingleFeatureDistanceSelectors~
use exclusively a
single-feature distance. To be precise, for each $v_j$, we create an
operator selector that computes the distance between \FState~A and \FState~B as 
$|A.v_j - B.v_j|$.

In addition to our simple single-feature approach, we consider the use of random 
projection \cite{DBLP:conf/focs/LarsenN17,DBLP:conf/kdd/BinghamM01}, a relatively 
straight-forward method of dimensionality reduction that has nice theoretical 
properties, good time-complexity, and which has found wide-spread application.
Further, in contrast to the selectors we detailed in the prior paragraph, random
projections produce their rankings based on multiple \FState-features.

First we produce \writtenNumRandProjectVecs~ random vectors in 
$\mathbb{R}^{\numStateAttributes}$
with unit Euclidean norm, generated by sampling each component uniformly at random on
$[0,1]$ then normalizing the result. Let $u_{p,i}$ be the $i^{th}$ such vector
produced (the $p$ subscript is to remind us this is for ``projection''). Let 
$\phi_{i}(x) : \mathbb{R} \rightarrow \mathbb{R},~i \in [\numStateAttributes]$,
be a strictly increasing function of $x$ --- we will specify what
these are and why we use them in just a moment. For a selector using
random projection with projection vector $u_{p,i}$ and featurization functions
$\phi_{j}$, the distance between \FState~A and \FState~B is computed as
\begin{equation}
\label{eq:randomProjectionDistanceComputation}
	|~~~\big( \sum_{j=1}^{\numStateAttributes}\ithval{u_{p,i}}{j}\phi_{j}(A.v_j)\big) - 
	\big( \sum_{j=1}^{\numStateAttributes}\ithval{u_{p,i}}{j}\phi_{j}(B.v_j)\big)~~~|
\end{equation}
or, equivalently, as 
$$|~\mathscr{L}2\big(~ u_{p,i} ~,~\seqNotation{\phi_{j}(A.v_j) - \phi_{j}(B.v_j)}{j \in [\numStateAttributes] }~\big)~|$$
where we use $\mathscr{L}2(\cdots,\cdots)$ to be the $L2$ inner-product.

Our featurization functions serve to put disparate fields  
on some common ground
so that it is sensible to compare or combine them. We consider two methods to do this:
the first is standardization, namely: 
\begin{equation} \label{eq:standardizationForRandomProjectionOperatorSelectors}
\phi_{j}(A.v_j) = \frac{A.v_j - mean(\{q.v_j | q \in Q\})}{ std(\{q.v_j | q \in Q\})} 
\end{equation}
Standardization causes each $A.v_j$ to have the same mean and standard deviation (zero
and one, respectively), but preserves distances between values, in the sense that,
for \FStates~ A, B, and C such that $A.v_j \not= B.v_j$, we have:
\begin{equation} \label{eq:randomProjectionSelectors:standardizationPreservesDistances}
\frac{A.v_j - C.v_j}{A.v_j - B.v_j} = 
\frac{\phi_{j}(A.v_j) - \phi_{j}(C.v_j)}{\phi_{j}(A.v_j) - \phi_{j}(B.v_j)}
\end{equation}

Our second approach is $\phi_{j}(A.v_j) = ECDF(A.v_j, \{q.v_j | q \in Q\})$ --- that is, in the 
distribution of $v_j$ values, this featurization gives the proportion of the distribution that  
have value no greater than $A.v_j$. \footnote{Notice that in both of the approaches, the transform is 
based on $Q$ as opposed to $Q_{f_t}$; the latter would have been preferred, but for a variety of reasons, 
our implementation does not at this time use $Q_{f_t}$ in those capacities.} 
Unlike standardization, the ECDF value has an a priori bounded image of $[0,1]$, and
random variables transformed by the ECDF\footnote{Also known as the 
probability integral transform } have very well understood behavior 
(using the universality of the uniform distribution and Dvoretzky–Kiefer–Wolfowitz inequality to uniformly 
bound the error between the true CDF and the ECDF).
The main distinction we draw between standardizing and using the ECDF, for our application,
is whether or not the "raw" feature distance (as given by standardization) is more meaningful than the "distance across the
distribution" (as given by taking the ECDF). In situations where the significance of a ``raw'' distance value varies
substantially with where that distance is based in the distribution, the ECDF may provide a more sensible
ranking of values. More broadly, ECDF can be used to measure distance in a way that is 
less effected by \textit{common} artefacts of  
data representations that may be chosen; specifically, given a collection
of data in one variable, applying a strictly increasing function to the data can radically change the distance between
a pair of points, but the ECDF will be unaffected.
Further, while the random projection method we adopt is neither a copula nor a difference between copulas, 
aspects of what make copulas attractive for modeling similarly motivate our use of ECDFs.
Ultimately, whatever the case, we'd like to use whichever of the distribution's properties best correspond with our ``class label'' 
--- the operator(s) that is/are best to apply to the \FState~in order to  
move the description's abstraction level sufficiently in the proper direction (which, as an effect, 
should satisfy a request of type $f_t$). In the next subsection
(\cref{subsubsec:firstOrderSelector:computingVoteFromRanking}), we go over how we leverage previously seen ``labels''.

For efficiency, both in terms of placement in the memory hierarchy and time complexity at scale, we compute the ECDF
using reservoir sampling. The mean and std are computed efficiently by tracking the sum, squared sum, and
number of entries for each field.

\subsubsection{Computing Vote Distributions Given
\CapFState-Distances}
\label{subsubsec:firstOrderSelector:computingVoteFromRanking}
Let $S' \in S_s$ be an arbitrary history-informed selector, and let $\alpha$ be a fixed value
in $(0, 1)$ specified ahead of time for use by $S'$. Let
\begin{equation}
\label{eq:weighingSuccessByDistance}
\text{weight'}(O') = \sum_{q_t \in 
	Q_{f_{t}}}y_t\mathbbm{1}(O' = O_{t})\alpha^{\textbf{rank}(q_t)}
\end{equation}
where $O' \in S_O$. The vote distribution returned by $S'$ is produced by running $\text{weight'}$
over each element of $S_O$, then normalizing the resulting list of values. 
Notice that the above description explicitly uses operator success rates and notions of
distances between \FStates, covering both 
\ref{considerationA2} and \ref{considerationA3}.
Further, to a very limited extent, the filtering used to form $Q_{f_{t}}$
in \cref{subsubsec:firstOrderSelector:rankingAndVoting}
aids in pursuit of consideration \ref{considerationB2}.
For each field considered, we create two selectors, one whose
value for $\alpha$  is $0.811$ and one
whose value for $\alpha$ is $0.896$. Assuming there are no ties and that the
use-history is sufficiently long, an $\alpha$ value of $0.811$ places roughly
90\% of the mass in the top 10 positions, while an $\alpha$ value of $0.896$
places roughly 90\% of the mass in the top 20 positions.

While many reasonable alternatives for weighing operators would incorporate
information about
how often an operator has been used (i.e.,\cref{considerationA1}) in addition to its success rate
(i.e., \cref{considerationA2}), we limit ourselves here to
considering only the latter in part because
the final step in selecting an operator incorporates the former via the 
UCB algorithm.

An alternative that was  implemented, though currently deactivated in our code,
is the utilization of
KNN-like approaches for weighing.\footnote{KNN: K-Nearest Neighbors (``K'' here being unrelated to variables that appear 
elsewhere in this document)} For these selectors, we limit
consideration to the top $z$-neighboring \FStates~
(for some integer $z$ that is a fixed parameter of the selector) and vote for
each operator in proportion to its success rate
over these \FStates, providing no voting mass to other operators.
KNN approaches can be implemented 
efficiently (especially when over a single variable\footnote{In multiple 
dimensions, a KD-tree (e.g., \cite{DBLP:journals/cacm/Bentley75,DBLP:journals/algorithmica/Sproull91} ) allows for efficient search for nearest
neighbors, albeit approximate. In a single dimension, a sorted list trivially
allows efficient and exact answers.}), and offer a different profile of biases
compared to the exponentially weighting (e.g., how they treat operators that are frequently
used, but only for \FStates~ distant from the current one).
While interesting to consider, we have opted against using this KNN-like method within our current implementation
due to unencouraging  empirical observations thus far, suggesting that the approach does not sufficiently leverage
available information while attempting to utilize the distance score.
Our exponential weighing scheme seems capable of leveraging all pertinent
information, and arguably utilizes distance in a more natural manner.
However, we might reverse this decision at a later time, after further experimentation with appropriate resources.

\subsection{Higher Order Selectors}
\label{subsec:higherOrderSelectors}

Via the infrastructure described in \cref{subsec:inference:furtherFeaturization},
we support modeling joint-behavior between 
selectors by constructing 
a new selector that acts as a function of other selectors' output.
Among the benefits this provides, it enables
combining the opinions of multiple selectors that each examine independent information streams.
In total, \numberProductSelectors~ out of  the \numberOfSelectors~selectors we have implemented
fall into this category.

In our implementation, we produce higher-order selectors
whose votes are the normalized product of two history-informed and/or applicability
selectors;
in the case where the support of the product is empty, the uniform
distribution is returned. 
For each pair of selectors that vote based on the single-feature distance of distinct
fields,
we form a unique new selector.  Additionally,
we form a unique new selector for each pairing of a history-informed selector\footnote{I.e., both
those based on single-feature distance and those using random projection} and an
applicability selector.

\section{Experiments}
\label{sec:experiments}

Our experiments are on-going. Currently, we using two simple domains --- the testing domains, as 
described at \cite{david_bayani_2021_5513079}  
--- as fast and generic arenas that we 
understand well, and thus allow us to better interpret and extend results. In contrast,
if we used complex domains and learned systems from real-world scenarios for our
initial
evaluations, not only would it require greater resources, it would introduce numerous confounding factors
and make it more difficult to attribute outcomes to causes. Further, we would like to be able to 
learn lessons from one domain, make improvements, then be able to evaluate on a new domain, to help ensure
that any attempted modifications were not ``overfit'' by the experimenters to the characteristics of one
domain. 

Currently, trials are produced on an on-going basis by having questions randomly generated for the target domains (similar
to the generation process described in section 4.2 of \cite{https://doi.org/10.48550/arxiv.2006.12453}). We bound the length of time
each question-response session may run, as well as the amount of memory allowed to Fanoos; if these bounds are exceeded, our
experiment harness forces the sessions to exit.

After an extended period of running our process, we plan
to examine:
\begin{enumerate}[{label=(R\arabic*)}] 
\item{ Which selectors receive the largest (or smallest) magnitude weight, and  whether the 
	weight is positive or negative. Discussing the distribution of weights also may be
	of interest.}
\item{ Which operators were used most often. Also worth examination here are the weights given to the uninformed selectors that 
	vote exclusively for one particular operator (see \cref{subsec:uninformedAndApplicabilitySelectors}),
	since the occurrence of related activities are reflected in the weights of those selectors
       (those activities being the number of times its targeted
	operator was not used, the number of times the operator was used and succeeded, and the number of 
	times the operated failed when used).}
\item{How quickly Fanoos --- using the learning method described in this paper with the reward function (i.e., \autouserShort)
    prescribed --- learned which operators are appropriate  to use in different circumstances.
    }
\item{\label{enum:item:plotOfSuccessRate} The running success rate across time (including bounds to account for discretization effects on the value)}
\item{Distribution statistics (e.g., mean and variance) for the 
    change in pertinent measures that occur after application of the blank operator. This provides a baseline
    to judge the performance of other operators and evaluate other measures.  
    }
\item{ Analysis based on human-provided word-labels, similar
    to section 4.3.3 (``Human Word Labeling'') of \cite{https://doi.org/10.48550/arxiv.2006.12453}.
    The labels are seen neither by Fanoos nor by \autouserShort, allowing us to use the information as a separate evaluation
    for both with minimal explicit contamination.}
\item{Qualitative analysis of descriptions (similar to the experiments in \cite{https://doi.org/10.48550/arxiv.2006.12453}) 
    generated under different regimes of training and at different points in time.}
\item{ If space, resources, and interest are amenable to it, case-studies of interactions  
    taken from different durations into the
    learning process (e.g., at the beginning, after UCB bounds become ``small'', mid-way before performance
    plateaus, and after performance stabilizes).
}
\item{Examination of the final vote distributions (i.e., \cref{eq:inference:DSample} ) at each step. For example, we may
    plot the per-timestep entropy and the time-averaged entropy (the latter with appropriate bounds).
    We expect the entropy of the final vote distribution to generally decrease from initial values as 
    learning progresses. We also expect the time-averaged entropy will  plateau
    at a lower-bound eventually, presumably around the time of when roughly
    maximum performance is reach; we would take this to indicate that Fanoos has developed stronger 
    ideas/opinions as to a what operators are worthwhile to apply in various circumstances. 
    }
\item{\label{enum:item:strengthOfAutouserOpinion} 
    A plot over time of the ``strength'' of \autouserShort's opinion on a subject, as shown by the value of:
                $$\frac{\frac{S_1}{S_2} - g^{\ell}}{g - g^{\ell}}$$
    where the variables in question come from \cref{algo:autoUserDeterminesFeedbackToGive}. Note that this
    value is negative when \autouserShort~ will reissue the same request, greater than or equal to one when the \autouserShort~
    issues the opposite request, and is in $[0,1)$ when the response provided by \autouserShort~ is influence by
    randomization. Both raw values and the time average with variance bounds may be shown. 
    Comparison to the plot of success rate over time (\cref{enum:item:plotOfSuccessRate}) may aid 
    interpretation --- namely that as $g$ increases, the strength of \autouserShort's
    opinion should on average be lower.}
\item{Analysis of the correlation between final vote distributions in pairs of consecutive timesteps, $t$ and $t+1$. 
    Consideration of this topic may benefit from (a) further stratifying analysis buckets by whether the timesteps 
    represent a successful adjustment (i.e, by whether $f_{t} \not= f_{t+1}$) or (b) a scatter plot of the correlations versus
    the ``strength or the \autouserShort's opinion'' (as defined in \cref{enum:item:strengthOfAutouserOpinion}) 
    at timestep $t+1$.\footnote{
    Recall from \cref{algo:autoUserDeterminesFeedbackToGive} that randomization plays a role in determining the 
    final feedback that \autouserShort~provides, thus (b) gives a sense of how much or how little the feedback was 
    based on chance, in contrast to (a).}}
\item{Resources and interest permitting, ablation studies as state-fields used by the \autouser and/or selectors are removed and/or added.
    We already provided some comments regarding the utility of these experiments in \cref{subsec:autouser:method}. }
\item{Resources and interest permitting, examination of behavior when selectors and/or operators are added or removed,
    whether it be at the beginning of the learning process or after some time into it.
    In the case of selectors, it would make sense
    to categorize them by their weight (strong
    negative, strong positive, near zero, or other) and show the behavior when each category is removed. 
    Motivation for these trials come not only from the desire to build a scientific understanding, but also from 
    engineering considerations, such the desire to enable and understand the ramifications of  
    adding novel innovations in a hot-swap fashion.
    }
\item{Resources and interest permitting, examination of the overall effects of interpreting user-invoked exiting of the description
	adjustment loop in the fashion described in \cref{subsec:learningProducingTheRewardSignal}. Our comments at the end of
	\cref{subsec:autouser:method} are relevant here. Experiments under this heading could include performance analysis
	after the reward for  $f_{t}=$``b'' in \cref{tab:rewardFunction} is changed to 0 and/or 1, in addition to other analysis of
	Fanoos's sensitivity to the value of $\varForNumberBadRefinementsAutouserIsAllowed$ used by the \autouser.} 
\item{Sanity checks on the frequency that cycles (if any) occur. This would also include analysis of the
    distribution of cycles' periods and content. Comments regarding cycles in
    \cref{subsec:autouser:method} and \cref{appendix:ourJudgementOfCycles} pertain here. }
\end{enumerate}
After demonstrating on the test models and gaining insights from those trials, naturally we also 
aim to showcase on real models, performing similar, rigorous analysis to the extent possible and practical. 

Ideally --- resources providing --- user studies will also be conducted to demonstrate:
\begin{itemize}
\item{Improvements garnered by the learning process (and therefore also the \autouserShort), 
     as shown by human blind evaluations on before-learning and after-learning comparisons}
\item{Measurements of human satisfaction with the adjustments provided by the system 
     after it has been trained to the point of plateaued reward under \autouserShort .} 
\item{Examination of whether \autouserShort~ causes the average success rate of Fanoos, after fine-tuning with a  
    human, to be as high or higher than if Fanoos was trained from scratch by a human}
\item{Examination of whether the time required to 
    fine-tune Fanoos for humans to perform satisfactorily or maximally after training with \autouserShort~
    is substantially less than the time required for a human to train Fanoos from scratch to achieve 
    similar levels of  stable, average performance. 
}
\end{itemize}

\section{Further Extensions}

Comments regarding possible extensions of this work that may be of interest
for readers to consider can be found in \cref{sec:moreFutureWorks}.

\bigskip
\noindent\textbf{Acknowledgments.}
We would like to thank Dr. Stefan Mitsch\orcidID{0000-0002-3194-9759} for his
kindness, supportiveness, and  
interest during this venture.

\bibliographystyle{splncs04}
\bibliography{Bibliography-File}

\appendix

\section{Several Interpretations of the Operator Selection Process}
\label{appendix:interpretationsOfOpSelectionProcess}

The method used to segment blocks of selectors can be compared to a variety
of techniques from ML and rule-based systems. We have already exposed
how the connection to featurization in  generic ML is relevant; by having
features
that are non-linear functions of the base-set of selectors, we can easily
form non-linear decision boundaries even if the final aggregation method is
linear in respect to the features --- this is common practice in generic ML, for instance, while
using SVMs.

Beyond seeing selectors as barely informed , observable aspects of
phenomena --- as basically just data without deeper processing ---
we may also  view them as experts, in the sense often used
to motivate randomized weighted majority voting (\cite{littlestone1994weighted}) or other simple
forms of zero-regret learning.
While easily interchanged from a purely mathematical outlook,
this latter perspective of the inputs provides a different sensibility with which
to view the selectors. Similarly, considering the approach from the standpoint
of literature for general ensemble-methods 
\footnote{
While we have been aware of general connections to ensemble methods such as bagging, boosting, cascades etc., 
we very recently came upon the lead to stacked methods (\cite{DBLP:journals/nn/Wolpert92,DBLP:journals/ml/Breiman96a})
that use combiners based on linear models. We are still in the process of adequately exploring this particular
connection, but early indications suggest it is a fruitful tie. } 
%
provides still a slightly
different lens to view the process and, worth emphasizing, yet another tool-bag of tricks that
may be applicable. Motivated by the desire to perform better than the single best selectors, 
we suspect the ensemble viewpoint would lead down more roads of benefit than placing concerns
over minimizing regret in the front of mind.\footnote{Notice that in the preceding 
discussion, we were commenting on lenses to view the inputs of the system, not necessarily
commenting on additional desiderata we want to adopt.}

Outside of ML-based approaches,
literature on scalable rule-based system architectures provide another perspective
for comparison, particularly those based on blackboard architectures
\cite{DBLP:journals/ai/Hayes-Roth85,10.1145/356810.356816,DBLP:journals/simpa/Straub22}.

An immediate question that may spring to the reader's mind is what utility, if anything,
pointing to these connections provides. Let us first note that,
regardless of the perspective taken, all ways of viewing our work would be in regard to the 
same concrete facts --- namely, our work. As tautological as this sounds, the distinction being 
drawn is similar if not the same as the difference between a particular mathematical structure (
in the sense the work ``structure'' is used in classical logic), and various logics
that may consider \textit{that} structure, but under differing rules and syntax. That is,
while all the mentioned literature-lenses consider the same concrete object, they provide
differing evaluation criteria, either explicitly or implicitly via the history, trends, 
social-structure, and connotations of the area. Of particular interest to a problem solver (e.g., 
the researcher or engineer trying to extend this work), these different perspectives provide
different paths and (implicitly if not explicitly) heuristics from which to work off of.
For those interested in a deeper discussion of the almost ``meta-research'' notions presented here,
the beginning chapters of \cite{10.5555/86564} provide a reasonable starting place.

\section{Further Comment on How We Currently View Any Cycles that Occur in the Fanoos-\Autouser~Interactions}
\label{appendix:ourJudgementOfCycles}

Arguably, not all types of cycle are bad. For instance, if Fanoos produces a repeating series of descriptions in response
to  similarly repeating requests by the \autouser, so long as \autouserShort~ is being satisfied and the cycle has a
period of considerable length, this may well help strengthen Fanoos's knowledge of how to satisfy requests on cases that
may be particularly frequent. While we admit that the possibility of the \autouser~ (as we currently implement it)
being involved in cycles is a point of departure from ideal human behavior (namely, searching through
descriptions like a binary tree, and using the special, user-invocable history-travel operator\footnote{See 
``u'' in \cref{tab:typesOfUserFeedback}.} in case they wish to
return to an earlier state / description), the benefits just mentioned may still be present and might outweigh any detrimental
effects. This all said, if this is ultimately deemed a problem, it would be 
trivial to modify the \autouser~ to issue punishments to Fanoos if the latter ever repeated a description;
this in-and-of-itself, however, would not prevent future cycles from occurring and may cause 
Fanoos to ``become confused'' as to the reason it is receiving negative feedback. Risk for ``confusion''
may be particularly high if none of Fanoos's
selectors are sufficiently sensitive to the structure of the interaction history --- essentially, 
how could Fanoos know it should not repeat descriptions if it lacks ``long-enough-term memory'' to know it 
repeated itself? It may be possible to detect a cycle via latent factors in the current timestep's state,\footnote{Bear in 
mind that a state need not repeat in order for a description to repeat. In fact, taken fully, states cannot
repeat since each contains the history of interaction --- however, it is possible that all other variables
in a state at timestep $t$ match those of some prior timestep.} but that seems doubtful and at best 
not sufficiently reliable. 
Naturally, a way to 
help address these concerns is to either modify existing selectors or add new ones that incorporate this
additional knowledge, voting in a way that aim to avoid cycles while also pursuing the other pertinent objectives.

\section{Explanation and Derivation of How the Total Number of Boxes are Used in \cref{subsubsec:firstOrderSelector:determiningDistance}}
\label{appendix:derivationOfHowNumberOfBoxesAreUsed}

In our refinement scheme, per the description in \cite{https://doi.org/10.48550/arxiv.2006.12453},
we bisect or trisect the axes of the input boxes until no pertinent axis\footnote{Recall from \cref{subsec:parameterAdjustingOperators}
that the operators may change the set of axes that are candidates for refinement in some circumstances.}
is longer than $\epsilon$. Under such operation, that makes the maximum number of boxes:
$$3^{\zeta_{I}\lfloor log_3(\epsilon)\rfloor}$$

Under a typical normalization, then, we would have:
$$\frac{\text{total\_number\_boxes}}{3^{\zeta_{I}\lfloor log_3(\epsilon)\rfloor}}$$
However, we use a logarithm of the above, which, after trivial simplification, gives the 
formula shown in \cref{subsubsec:firstOrderSelector:determiningDistance}:
$$log_{3}\Big( \frac{\text{total\_number\_boxes}}{3^{\zeta_{I}\lfloor log_3(\epsilon)\rfloor}} \Big)$$
We were motivated to use a logarithm first and foremost by what would make sense for the 
distance measure:  
given how our refinement scheme works and the multi-dimensional nature of our abstract states,
the proportional change in the number of boxes is typically more significant than the raw change
in number. This, for instance, reflects the intuition
that the impact of having $b_1$-many boxes versus $b_1 + c$ for some constant $c$ seems to 
decrease as $b_1$ increases --- especially given that our choice of bisection versus trisection
is random.
In additional to this 
rationale, generally speaking we gain numerical benefits 
from using a logarithm in this capacity, and the resulting equation is simpler.

\section{Other Future Works and Future Additions}

\label{sec:moreFutureWorks}

Fanoos and the extensions of it detailed in this paper are amenable to the addition
of numerous components, features, and improvements. In this work, we have
described a sensible and effective method of allowing Fanoos to learn  to select appropriate
operators from a large collection of options in order to respond to a user's request for greater or lesser
abstraction of a particular state's description. While this work can be even further extended
in interesting ways, we (almost by necessity) leave a number of such promising items as future work.
In the body of this paper, we have already highlighted some avenues for further improvements and capabilities;
here, we provide a small sample of additional add-ons that did not fit into the main body or would
be too far distracting if included there.
Listed in not particular order:\footnote{The text ``FE'' that appears next to the numbering stands for
``Further Extensions''.} 

\begin{enumerate}[{label=\tiny{\textbf{FE}\arabic*}}]
\item{Statistics about the abstract states that reflect their spatial distribution in a way that is
not tied to one specific domain. Such information would help further inform selectors. Potential
statistics include summary values for (a) the distribution of radii between the box-centers and the 
center of the universal bounding box or (b) the distribution of distances between pairs of box-centers
(possibly found through random sampling, for the sake of efficiency).}
%
%
\item{For use by selectors when dealing with abstract states over the output space:
the addition of statistics derived from the output-box volumes normalized by 
the approximate image of the learned system. The approximate image for the learned system
would be found by pushing the input-space's universal bounding box
through the learned system via our abstract domain analysis.}
%
%
\item{ Possible modification to the reward function to include an exponential decay based 
on the amount of time it took to produce a result --- for example, including a term such as
the value $\text{exp}(-(\text{time\_taken} - 60\text{minutes}))$ normalized between zero and one.
This would help promote generation of faster descriptions --- balanced, of course, with other 
performance criteria.}
%
%
\item{ Addition of selectors that leverage knowledge of the operator's internal structure.
The motivation for this is also reflected in \cref{considerationA3}, but in this work, we have primarily
focused on leveraging similarity between states. Our framework immediately facilitates the
addition of such information about operators. The operators
do have internal structures that selectors in Fanoos could in principle access; in the case of parameter
adjusting operators (\cref{subsec:parameterAdjustingOperators}), computing some sort of similarity between
operators should be straight-forward. For instance, while we're not necessarily advocating for use of a 
Euclidean distance here, it is the case that such a metric could be directly used for parameter adjusting operators. \\

One can easily brainstorm ways with which selectors can aggregate 
over similarity between operators after an initial aggregation over states in order to produce a final 
vote distribution including both knowledge sources. Even more simply, as a starting place,
one can leverage the infrastructure established for the applicability-based selectors 
(\cref{subsec:uninformedAndApplicabilitySelectors}): for instance, a selector can be made that 
adds more weight to operators that re-form boxes with larger refinement parameters
when the request is ``m'', and does the opposite when the request is ``l''. 
In addition to home-brewed methods, literature 
pertaining to action selection in continuous action spaces may yield some insight  (such spaces often have
natural notions of distances between their members), though caution must be taken since our setting lacks
most arithmetic properties that are useful in continuous settings (e.g., \textit{generally speaking}, it would not
be  possible to ``average'' over our operators --- with current arrangements, even if such arithmetic was defined, 
our set of operators would likely not be closed under it).
}
%
%
\item{Methods for learning to apply multiple operators in a row prior to querying a user for further feedback (perhaps
lumping the multiple basic operators into one ``meta-operator'').}
%
%
\item{The addition of predicate constraining operators (\cref{subsec:predicateConstrainingOperators}) that 
select predicates based on information regarding the volume that the predicate covers.}
%
%
\item{ Simple but not yet included in the standard library for facilitating generation of predicates: the inclusion of a 
negation sign on predicates, operating using the similar framework as our conjunction to act on predicates. Negated predicates
can easily be included as though they are primitive predicates
(i.e., from a mechanical standpoint, they could be handled the same way when forming descriptions). A tweak that might
be desired if this is pursued (other than trying to leverage what we know to boost efficiency compared a simple
implementation) is to add a heuristic that prefers non-negated predicates over negated predicates when making final decisions.
Such a heuristic would likely be unneeded in most circumstances, however, by nature of how we filter for predicates
that are sufficiently specific to a box (that is, in most circumstances where there are candidate literals that 
are not negated, we expected a negated predicate to not be among those that are most specific to a box);
see algorithm 1 in \cite{https://doi.org/10.48550/arxiv.2006.12453}.
}
%
%
%
%
\item{ Operators and facilitating infrastructure in our refinement process allowing users to select predicates in a 
description to be replaced with different description content, the latter generated via changing the refinement of the
chosen predicate's underlying abstract states. This proposal is in contrast to the predicate constraining operators
in \cref{subsec:predicateConstrainingOperators} which work on the syntactic level, disallowing or reallowing use of
a symbol. The process to incorporate this proposal's feedback would be to find those boxes consistent with the 
sub-condition (predicate, conjunct, a subset of disjuncts, etc.), sufficiently refine those boxes more/less, then 
ideally leave the rest of the description generation process alone.
An example of the process envisioned: suppose the user asks a question $quest_A$ and Fanoos says that such a thing occurs 
when conditions $B$, $C$, or $D$ happen. The user then may ask for a more concrete description of the occurrences that $B$
refers to. Fanoos then refines the boxes consistent with $quest_A \land B$ further and does not modify abstract states
that are outside that collection (such as those consistent with $quest_A \land \lnot B$). }
%
%
\item{Additional experiments that demonstrate Fanoos elaborating the behavioral differences between over-, under-, and correctly
trained policies that operate in the same domain.}
%
%
\item{Per discussion in \cref{appendix:ourJudgementOfCycles}, modifying Fanoos and / or the \autouser~ to 
prevent and/or be punished for repeating a description during a single description adjustment loop --- if, that is,
such a modification is ultimately determined to be desirable.}
%
%
\item{For the \autouser, those curious might like to look into dynamically changing the cutoffs used in \cref{eq:vectorPsi}  
    for $j\in\gamma_2$, motivated by similar reasoning as the dynamic threshold used in 
    \cref{algo:autoUserDeterminesFeedbackToGive}.}
%
%
\item{ An additional selector that functions similarly to the random projection (see \cref{subsubsec:firstOrderSelector:rankingAndVoting})
but whose vector undergoes a perceptron-like updated based on feedback. That is, if $s_i$ is the selector and
the operator chosen at the end of the process is $O_t$, then the reward signal used in a perceptron-update of $s_i$'s projection 
vector is 
$$y_t(~~\mathbbm{1}\big(~(s_i(*_t))(O_t) > U\big) - \mathbbm{1}\big(~(s_i(*_t))(O_t) < U\big)~~)$$
(see \cref{subsec:learning:updatingSelectorWeights} for definition of terms).\footnote{
The reward signal is intentionally zero when $(s_i(*_t))(O_t) = U$. It is perhaps possible
that such an arrangement could cause the projection vector to get stuck at an undesirable value
in pathological cases (e.g., the projection vector becomes zero), but given how we divy votes
after establishing the distances, it is very unlikely that a ``stuck-position'' would be stable over the
long-term if it ever were to occur. That is, in pathological cases the projection vector may spend
more time in the neighborhood of ``bad'' values than we'd like, but eventually the vector would
move away from it with great likelihood.
} }

\end{enumerate}

\end{document}